\pdfoutput=1

\documentclass[11pt]{article}

\usepackage[]{acl}

\usepackage{times}
\usepackage{latexsym}

\usepackage[T1]{fontenc}

\usepackage[utf8]{inputenc}

\usepackage{microtype}

\usepackage{subcaption}
\usepackage{times}
\usepackage{latexsym}
\usepackage{graphicx}

\usepackage{booktabs}
\usepackage{longtable}
\usepackage[T1]{fontenc}

\usepackage[utf8]{inputenc}
\usepackage[load-configurations=version-1]{siunitx} 
 \usepackage{listings}
\usepackage{microtype}

\usepackage{inconsolata}
\usepackage{multirow}
\usepackage[most]{tcolorbox}
\usepackage{booktabs}
\newcommand*{\thead}[1]{%
\multicolumn{1}{c}{\bfseries\begin{tabular}{@{}c@{}}#1\end{tabular}}}
\title{Analyzing LLM Behavior in Dialogue Summarization: \\Unveiling Circumstantial Hallucination Trends}

\author{Sanjana Ramprasad$^\diamondsuit$ \quad Elisa Ferracane$^\clubsuit$ \quad  \textbf{Zachary C. Lipton}$^{\clubsuit}$ 
\\
$^\diamondsuit$Northeastern University \\
$^\clubsuit$ Abridge AI \\
\texttt{\small \{ramprasad.sa\}@northeastern.edu} \\
\texttt{\small \{elisa,zack\}@abridge.com}}
\begin{document}
\maketitle
\begin{abstract}
Recent advancements in large language models (LLMs) 
have considerably advanced the capabilities of summarization systems.
However, they continue to face concerns about \emph{hallucination}. 
While prior work has evaluated LLMs extensively in news domains, 
most evaluation of dialogue summarization has focused on BART-based models,
leaving a gap in our understanding of their faithfulness.
Our work benchmarks the faithfulness of LLMs for dialogue summarization,
using human annotations and focusing 
on identifying and categorizing span-level inconsistencies.
Specifically, we focus on two prominent LLMs: GPT-4 and Alpaca-13B.
Our evaluation reveals subtleties as to what constitutes a hallucination:
LLMs often generate plausible inferences,
supported by circumstantial evidence in the conversation,
that lack direct evidence, a pattern that is less prevalent in older models.
We propose a refined taxonomy of errors, 
coining the category of "Circumstantial Inference"
to bucket these LLM behaviors.
Using our taxonomy, we compare the behavioral differences
between LLMs and older fine-tuned models. 
Additionally, we systematically assess the efficacy 
of automatic error detection methods on LLM summaries 
and find that they struggle to detect these nuanced errors. 
To address this, we introduce two prompt-based approaches 
for fine-grained error detection that outperform existing metrics, 
particularly for identifying "Circumstantial Inference." \footnote{The dataset can be downloaded from \url{https://github.com/sanjanaramprasad/circumstantial_inference.git}}

\end{abstract}

\section{Introduction}

\begin{figure}[t!]
    \centering
    \includegraphics[scale=0.5]{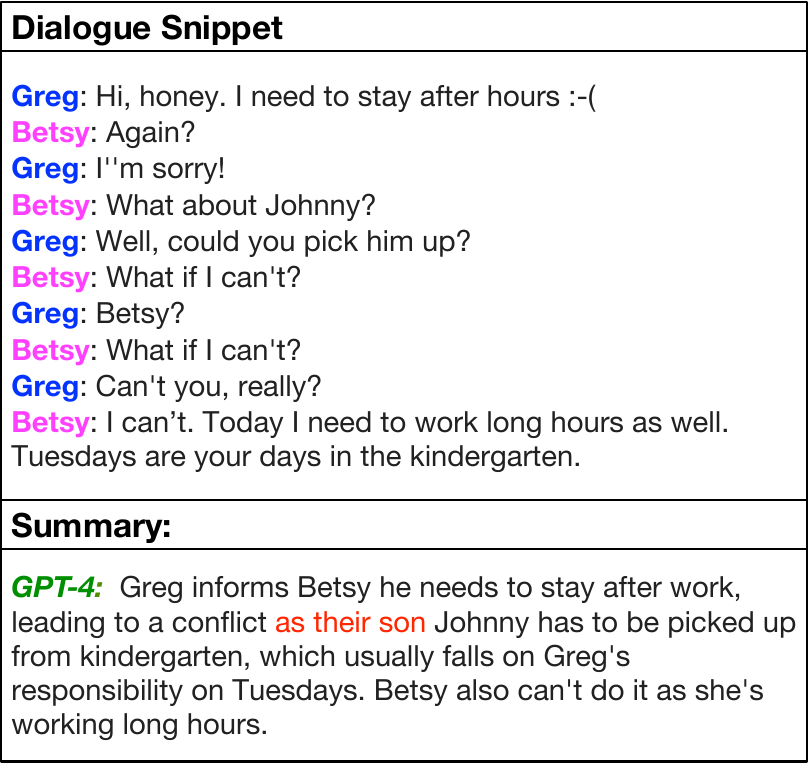}
    \caption{In the example provided, GPT-4 infers that the speakers are discussing "their son." Although this inference seems plausible given the circumstantial evidence in the conversation, it lacks direct evidence.}
    \label{fig:contextual_inference}
\end{figure}

Considerable progress has been made in summarization using large language models (LLMs) \citep{goyal2022news, zhang2023benchmarking}. 
However, the challenge of so-called ``hallucinations'',
characterized in this context as statements in summaries
that do not have direct evidence in the source material persists. 
As a result, evaluation of these summaries is an active area of research. 

In prior research, news articles have been the main testbed 
for LLM-generated summary evaluation \citep{zhang2023benchmarking, yang2023exploring}. 
Dialogue summarization remain less explored,
with prior works mostly focused on smaller fine-tuned models 
\citep{zhu2023annotating, gao2023reference, wang2022analyzing}. 
In this work, we close the evaluation gap,
focusing our analysis on LLM summaries of chit-chat style dialogues.
We obtain fine-grained inconsistency annotations 
for summaries generated (zero-shot) by two prominent LLMs
(GPT-4 \citep{luo2023chatgpt} and Alpaca-13B \citep{alpaca}) 
and across two summarization datasets 
(SAMSum \citet{gliwa2019samsum} and DialogSum \citet{chen2021dialogsum}).

In the domain of dialogues, a further gap exists 
in understanding the differences between 
summaries generated by LLMs 
and those generated by smaller fine-tuned models. 
In the news domain, prior work has found that LLM-generated summaries
have fewer inconsistencies \citep{goyal2022news, zhang2023benchmarking}. 
Work done by \citet{tang2022understanding}, also in the news domain, 
notes varying error distributions across different model categories. 
In our work in the dialogue domain, 
we compare differences in error rates and analyze the categories of errors 
for summaries of dialogues with fine-tuned models versus summaries with LLMs. 
As in the news domain, we find that LLM-generated summaries have fewer inconsistencies. 
Surprisingly, our analysis reveals that over 30\% 
of LLM-generated summaries contain inconsistencies, 
contrasting sharply with the inconsistency rate 
of less than 5\% in GPT-generated news summaries \citep{zhang2023benchmarking}.

    To further elucidate the differences between LLMs and fine-tuned models, 
    we annotate spans with error categories. 
    Previous work has primarily relied on part-of-speech-based tags 
    for error classification \cite{wang2022analyzing, zhu2023annotating, gao2023reference}. 
    However, complexities inherent in LLM-generated summaries, 
    often lengthier and more intricate, 
    do not neatly align with error categories based solely on part of speech, 
    warranting alternative strategies for a more meaningful categorization.
    Hence, our work proposes a refined taxonomy integrating existing error types. 
    We further introduce a \emph{new} error category specific to LLM behavior:
    \emph{"Circumstantial Inference."} 
    This category stems from the observation that LLMs frequently produce 
    statements that appear plausible based on circumstantial 
    (but not direct) evidence in the dialogues, an aspect hitherto unexplored.
    In particular, LLMs tend to produce statements 
    that may be \emph{circumstantially} implied 
    based on contextual cues in the conversation
    but not explicitly stated as seen in Figure \ref{fig:contextual_inference}. 
    Although these inferences are not directly stated and can be inherently unsupported, 
    they can still be useful in some instances,
    especially when summarizing ambiguous dialogues.
    However, the appropriateness of such inferred details 
    varies depending on context and domain, 
    highlighting the need for further investigation.
    
    In addition, there is limited understanding 
    regarding the automatic detection of the mentioned error types.
    Therefore, we systematically evaluate the performance 
    of state-of-the-art error detectors on LLM-generated dialogue summaries. 
    We also introduce two prompt-based methods for fine-grained error detection, 
    which notably outperform all prior state-of-the-art error detectors, 
    particularly in identifying the newly introduced error type, "Circumstantial Inference."\\
    
    In summary, our primary contributions are as follows:
    \begin{enumerate}
    \item We bridge a gap in understanding LLM effectiveness for dialogue summarization 
    by collecting fine-grained human annotations that highlight inconsistencies 
    and make the benchmark publicly available.
    \item We propose a refined taxonomy for error categorization of LLM-generated summaries,
    including a new error category called "Circumstantial Inference" 
    that captures the tendency of LLMs to produce plausible hallucinations based on conversation context.
    
    \item We examine differences in behavior in dialogue summarization 
    between LLMs and fine-tuned models by comparing error rates and types.
    
    \item We introduce two prompt-based methods for fine-grained error detection, 
    which notably outperform existing metrics. 
    These methods excel even in detecting the recently identified error type "Circumstantial Inference."
    Additionally, we evaluate state-of-the-art error detectors on model-generated summaries 
    across model categories and error types 
    unveiling their effectiveness and limitations.

    
    \end{enumerate}

\section{Human Evaluation: Zero-shot Prompted Dialogue Summaries}
\begin{table*}
\small
\centering
\begin{tabular}{p{1.7cm} >{\raggedright\arraybackslash}p{5.6cm} >{\raggedright\arraybackslash}p{7.3cm}}
\toprule
Linguistic Category & Summary Excerpt & Dialogue Excerpt\\
\midrule
Circumstantial Inference & Cameron is unable to bring a video game for {\color{red}their daughter} Peyton. & \multicolumn{1}{p{7.3cm}}{\raggedright Peyton: I have been asking you to bring that video game for me\\Cameron: Honey, I am not having enough time to come home.}\\
\\
Logical Error & Jane is worried about the travel time and suggests they {\color{red}meet later} &\multicolumn{1}{p{7.3cm}}{\raggedright Steven: the road is new, we will make it \\Jane: I don't want to stress out, let's meet at 4:30 instead of 5, ok?}\\
\\
World Knowledge & \#Person1\# plans to vote for {\color{red}Joe} Biden instead.&\multicolumn{1}{p{7.3cm}}{\raggedright\arraybackslash \#Person1\#: I will vote for Biden anyway.}\\
\\
Referential Error & {\color{red}Person1 said} that Person2 could call or email them. & \#Person2\#: Please call me or send e-mail.\\
\\
Figurative Misinterpretation & Alyssa {\color{red}likes} Fergie's national anthem. &\multicolumn{1}{p{7.3cm}} {\raggedright Alyssa: Have you seen Fergies national anthem?\\Derek: This is not normal. I saw it last week\\Alyssa: The best part is that she acts like she nailed it.} \\
\\
\bottomrule
\end{tabular}
\vspace{-0.7em}
\caption{Examples of linguistic categories for inconsistencies in {\color{red}red} between the LLM-generated summaries and the dialogues.
}
\label{tab:results_ling_categories}
\vspace{-1.5em}
\end{table*}
We aim to compare the difference in consistency 
of zero-shot prompted LLM-generated dialogue summaries 
with smaller fine-tuned model-generated summaries.\footnote{We 
deliberately choose zero-shot instead of few-shot 
to better understand the model's inherent capabilities.} 
To accomplish this, we conduct human annotations 
to identify inconsistent spans generated 
by both GPT-4 \citep{openai2023gpt4} and Alpaca-13b \citep{alpaca}. 
Specifically, we direct annotators to spot inconsistencies in summaries, 
marked by spans that lack evidence in the source text or distort information from it. 
Our evaluation is carried out on dialogue summarization datasets 
previously used for benchmarking fine-tuned summarization models.
\subsection{Datasets}
We perform human annotations on two prominent summarization datasets: 
SAMSum \citep{gliwa2019samsum} and DialogSum \citep{chen2021dialogsum}. 
SAMSum comprises artificially generated,
concise written conversations crafted by linguists, 
centering around everyday topics. 
Conversely, DialogSum presents a corpus 
of naturally occurring spoken dialogues 
reflecting real-life contexts.

To facilitate comparisons with earlier fine-tuned models, 
we annotate the same set of data points from previous benchmark studies, 
as outlined below:

a) Reference Matters (RefMatters): Introduced by \citealt{gao2023reference}, this dataset offers factual annotations for summaries generated on dialogues in SAMSum and DialogSum. The annotated summaries include outputs from four fine-tuned summarization models, addressing eight distinct types of factual errors: Entity, Predicate, Circumstance, Coreference, Discourse Link, Out of Article, Grammatical, and Others.

b) FacEval Dataset: Detailed by \citealt{wang2022analyzing}, this dataset provides annotations for BART-based models applied to the SAMSum dataset. It delineates six error types, namely Subject Object Error, Pronoun Error, Negation Error, Particulars Error, Hallucination Error, and Other Error. Notably, there exists a small overlap in data points with RefMatters.

\subsection{Models}
\label{sec:models}
We assess the performance of two prevalent Large Language Models (LLMs) in the context of dialogue dialogue summarization: (1) GPT-4 \citep{openai2023gpt4} utilizing the gpt-4-32k-0613 snapshot, and (2) Alpaca-13b  \citep{alpaca}. For both models, we use the default settings and prompt zero-shot using the following template to generate summaries:
\begin{tcolorbox}
Generate a summary of the following dialogue snippet: $\{\{Dialogue\}\}$
\end{tcolorbox}
Our evaluation compares our collected annotations against the inconsistency annotations from previous benchmarks. Specifically, we use annotations for BART \citep{lewis-etal-2020-bart}, UniLM \citep{unilm}, MV-BART \citep{mvbart}, and CODS \citep{cods} from the RefMatters Benchmark. Additionally, we consider annotations for BART \citep{lewis-etal-2020-bart}, MV-BART \citep{mvbart}, CondigSum-BART \citep{condigsum}, and Coref-BART \citep{corefbart} in the FacEval dataset. 

Henceforth, we refer to the above models as \textbf{FT-Summ}, representing smaller fine-tuned summarization models, and the zero-shot prompt-based models GPT-4 and Alpaca-13B as \textbf{LLM}.

\subsection{Fine-grained inconsistency annotation}
We perform two rounds of annotations to identify inconsistencies in dialogue summaries. 
\subsubsection*{Error Annotation}
In the first phase, we enlist two linguist fact-checkers from Upwork\footnote{\url{https://www.upwork.com/}} to assess summary consistency. Inconsistent summary sections are identified as those that conflict with or inaccurately represent dialogue information or lack sufficient evidence. This definition aligns with criteria from prior studies on news summarization \citep{polytype, xsumfaith, dae, hallucinated_but_factual}. Inter-annotator agreement, measured using pairwise F-1 metric, results in a substantial agreement of 66.94\%. Further annotation instructions are detailed in Appendix \ref{sec:annotation}.

\subsubsection*{Error Categorization}
In the second round of annotations, an author of the paper who is an expert linguist meticulously categorized the errors to attain more granularity. We find that traditional methods of error categorization that have relied heavily on part-of-speech tags \citep{wang2022analyzing, zhu2023annotating,gao2023reference} prove less effective when dealing with summaries generated by LLMs due to their tendency to exhibit increased abstraction and inference. This makes it challenging to align inconsistent spans with specific part-of-speech-based categories. Consequently, we propose a taxonomy of errors which we outline in the subsequent section.

\subsubsection*{Taxonomy of Errors}
\label{sec:taxonomy_err}
We describe the taxonomy of errors identified in this work below and provide examples in Table \ref{tab:results_ling_categories}. 

\textbf{Logical Error}: 
This category identifies inaccuracies in dialogue summaries. 
Our annotations highlight three main types of logical errors 
commonly found in summaries generated by LLMs:
a) Event misordering, where the summary presents 
an incorrect chronological sequence 
due to wrong word usage or sentence order.
b) Lack of commonsense, where models incorrectly reason 
through information that should be obvious.
c) Missed detail, where the summary would be correct 
if not for the omission of important information.
FT-Summ models also exhibit logical errors in this category, 
including inaccurate negations, wrong verbs, or incorrect word senses.

\begin{figure*}
    \centering
    \includegraphics[scale = 0.39]{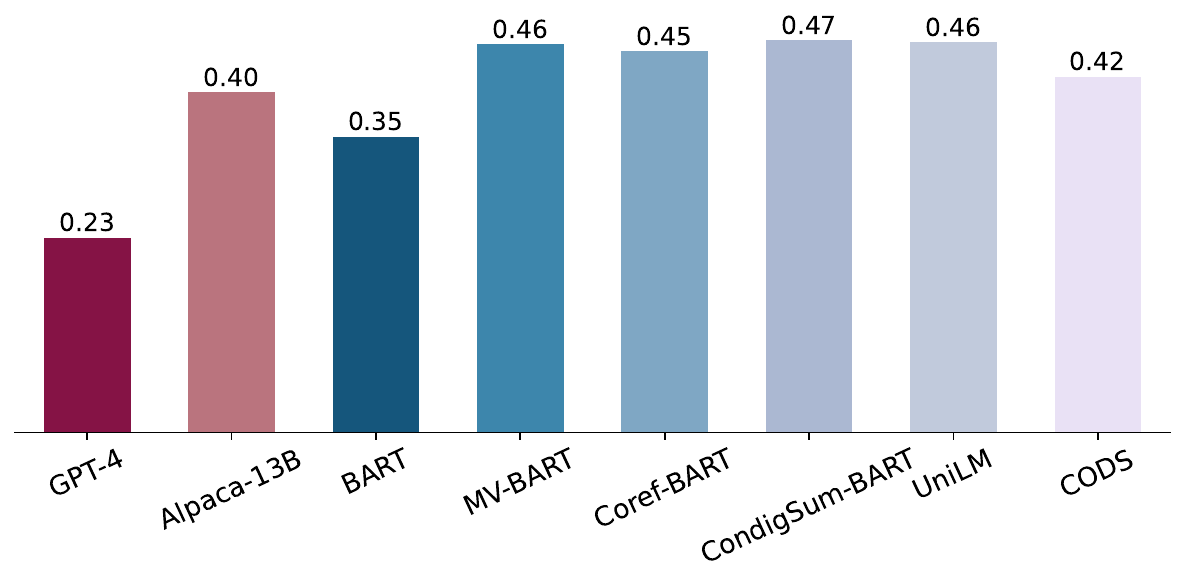}
    \caption{Each bar in this plot depicts the proportion of total summaries with inconsistencies across different model-generated summaries where GPT-4 performs the best (lower means fewer inconsistencies).}
    \label{fig:llms_vs_xformer_error_rate}
    \includegraphics[scale = 0.35]{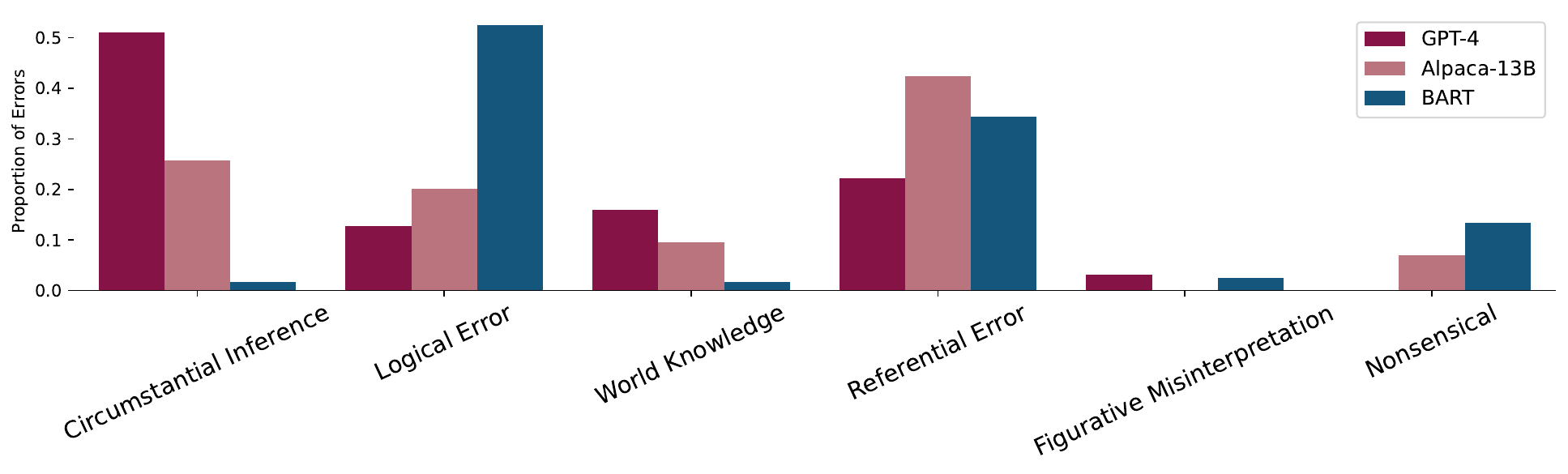}
    \caption{Error category proportions for each model in the dataset (lower values indicate fewer occurrences of specific categories). The more challenging circumstantial inference errors are common in GPT-4 but hardly present in BART.}
    \label{fig:llms_vs_xformer_coarse_error_categories}

    \label{fig:overall}
\end{figure*}

\textbf{Circumstantial Inference}: 
We introduce this new category not explored in prior work 
and inspired by Grice's Maxim of Conversation for Quantity, 
which states that cooperative speakers make contributions 
that are sufficiently but not overly informative \citep{grice1975logic}. 
Speakers intentionally omit information deemed shared knowledge. 
When the language model draws inferences based on circumstantial 
but not direct evidence in the conversation, 
we label this as a circumstantial inference error. 
While traditionally viewed as an inconsistency, 
we contend that in open-domain conversations, 
such circumstantial inferences may be reasonable. 
In the example listed in Table \ref{tab:results_ling_categories}, 
Cameron addresses Peyton as "Honey" plausibly 
because they both know Peyton is Cameron's daughter
(and explicitly stating that would violate the maxim of Quantity). 

The importance of these inconsistencies depends on the context. 
For example, in doctor-patient conversations, 
inferring a patient has diabetes from blood sugar discussions 
is more consequential than inferring general familial relationships.

\textbf{World Knowledge}: This error type constitutes 
a distinct subset of Out-of-Article Errors, 
wherein the inaccuracies involve real-world facts. 
For instance, the summary might include the full name of a public figure 
not explicitly mentioned in the dialogue (see Table \ref{tab:results_ling_categories}).

\textbf{Referential Errors}: 
This error resembles subject-object errors in prior research 
\citep{wang2022analyzing, gao2023reference}. 
However, LLMs show intricate misattributions, unlike FT-Summ models, 
where referential errors involve swapped entities. 
This complexity challenges previous straightforward categorizations.

\textbf{Figurative Misrepresentation} 
This category of error occurs when metaphors, sarcasm, or jokes in the content 
are mistaken for literal statements in summaries, altering the intended meaning.

\textbf{Nonsensical errors}: 
We consider grammatical errors in BART-generated summaries,
as well as instances where language models continue a prompt 
or repeat instructions after generating a summary as Nonsensical. 

\subsection{Evaluation Results}
\subsubsection*{Error Rates}
Figure \ref{fig:llms_vs_xformer_error_rate} depicts error rates 
for fine-tuned models (FT-Summ) and large language models (LLMs) 
applied to dialogue summarization datasets. 
Our results indicate that GPT-4 exhibits fewer inconsistencies 
in dialogue summarization compared to fine-tuned models.
However, this improvement is smaller than observed in prior research on news summarization, 
where GPT-3 achieved an error rate of less than 5\% \citep{zhang2023benchmarking}. 
Conversely, in our case approximately 23\% of GPT-4 summaries 
across all dialogue datasets display inconsistencies. 
Furthermore, Alpaca-generated summaries 
generally show lower inconsistency 
compared to most fine-tuned models,
but are surpassed by BART. 

On further analysis (shown in Appendix \ref{sec:error_rate_dataset}), 
we find that Alpaca outperforms the fine-tuned models (including BART) 
on the DialogSum benchmark \citep{gao2023reference}, 
but the opposite is true on the SAMSum benchmarks 
\citep{gao2023reference, wang2022analyzing}. 
Differences in Alpaca-generated summary quality across datasets 
may stem from variances in dialogue and summary features, 
to which larger language models may be less sensitive. 
DialogSum uses real spoken dialogues with multiple turns, 
potentially resembling pre-training data, 
while SAMSum involves synthetic written conversations. 

\subsubsection*{Fine-grained Error Category Distribution}
In the investigation of fine-grained categories, we conduct annotations on summaries generated by GPT-4 and Alpaca-13b for Large Language Model (LLM) models and BART within the framework of FT-Summ. We choose BART for annotation due to its consistent superiority in consistency compared to other fine-tuned summarization models, as evidenced by its performance across various datasets (see Figure \ref{fig:llms_vs_xformer_error_rate}).

One key discovery (shown in Figure \ref{fig:llms_vs_xformer_coarse_error_categories}) 
is the prevalence of Circumstantial Inferences in LLM-generated summaries. 
Roughly 38\% of errors fall into this category, 
which describes cases where assumptions are made 
based on circumstantial evidence within the conversation. 
Interestingly, these errors are rare in BART-generated summaries, 
constituting only 1\% of all errors. 

We also observe several error categories with lower prevalence in LLMs compared to FT-Summ. 
Notably, LLM summaries consistently lack grammatical errors, 
presenting coherent and well-written summaries.
However, we note a similar error type specific to LLMs---prompt errors. 
We group both LLM-based prompt errors and FT-Summ-based grammatical errors as "Nonsensical" 
in Fig \ref{fig:llms_vs_xformer_coarse_error_categories}. 
Interestingly, GPT-4 exhibits no instances of nonsensical text, 
whereas BART shows a higher prevalence of such cases compared to Alpaca.

Our analysis (shown in Fig \ref{fig:llms_vs_xformer_coarse_error_categories}) 
also uncovers a considerable reduction in logical errors,
specifically 17\%, in the case of LLM errors, in contrast to FT-Summ, 
where over 50\% of errors manifest as logical errors. 
This noticeable decrease signifies the superior proficiency 
of LLMs over FT-Summ models in deriving logical inferences from dialogues.

Despite advancements, persistent challenges in specific error types remain prevalent within LLMs. 
Both FT-Summ and LLMs exhibit a similar frequency of referential errors across all error categories. 
Notably, Alpaca-generated summaries show a higher prevalence of referential errors 
compared to BART-generated ones, elucidating the elevated error rate in Alpaca summaries. 
Further examination reveals two distinct types of referential errors: 
coreference and misattributions.
Coreference errors arise from the inability 
to accurately establish the reference of a pronoun or noun within a sentence, 
whereas misattributions involve erroneously attributing information or statements to the wrong source. 
Notably, in BART summaries, referential errors mainly consist of coreference errors (95\%), 
with misattributions constituting only 5\%. 
Conversely, referential errors in LLMs are characterized 
by a higher frequency of misattributions (58\%) 
compared to coreference errors (42\%).

\section{Automatic Error Detection}
Prior research has demonstrated a shift in error category distributions 
when summarizing with models of different generations, 
resulting in varied trends in the performance of automatic error detectors 
\citep{tang2022understanding}. 
This section aims to address the following key questions:

a) \textit{Do factuality metrics perform similarly 
on summaries generated by Large Language Models (LLMs) 
compared to older models (in our study, FT-Summ)?} 
Specifically, we investigate whether detecting factual errors 
in LLM-generated summaries poses greater challenges.

b) \textit{Which error types across various model categories
can factuality metrics identify, and what are the associated failure modes?} 
Drawing upon our error taxonomy, we analyze the ability of metrics 
to detect different error categories, with a specific focus on their performance 
in identifying circumstantial inferences, an aspect not previously evaluated.

c) Furthermore, \textit{we introduce a novel approach aimed at enhancing the detection of different error categories at the span-level} which we introduce in section \ref{subsection:Span_Detection}.

\subsection{Metrics}
To include a wider range of metrics, 
we assess metric performance in the following two specific contexts: binary classification (label the entire summary as factual or not) and span detection (identify the nonfactual span). 

\begin{figure}
    \centering
    \includegraphics[scale=0.61]{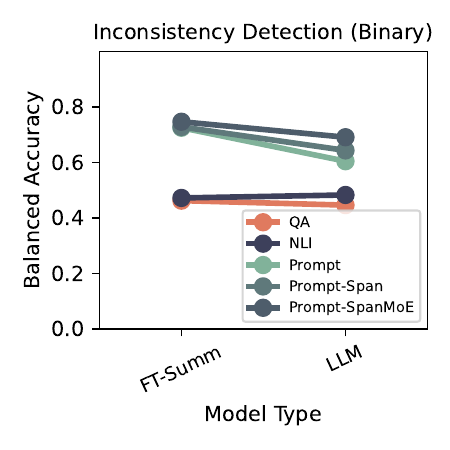}
    \caption{
Automatic error detectors exhibit varying performance 
when applied to FT-Summ versus LLM. 
While QA/NLI metrics indicate a slight improvement, 
prompt-based metrics are better in detecting inconsistencies 
generated by the FT-Summ model in comparison to LLMs.}
    \label{fig:binary_classification_xformer_vs_llm}
\end{figure}
\begin{figure*}
    \centering
    \includegraphics[scale=0.38]{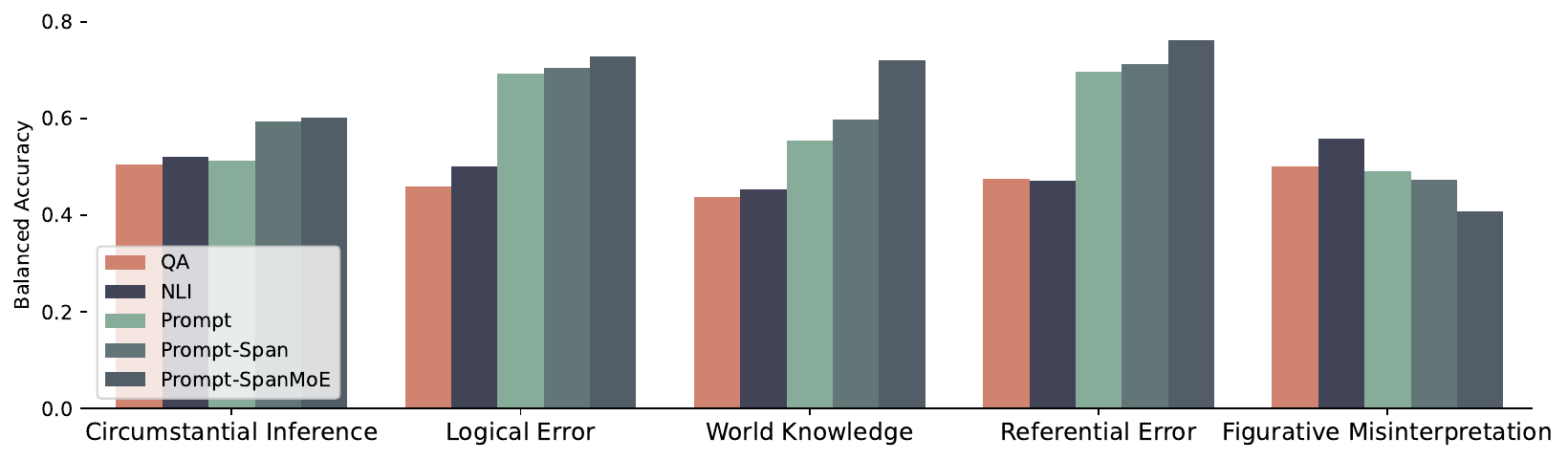}
    \caption{Inconsistency binary classification per error category}
    \label{fig:error_category_graph_binary}

    \centering
    \includegraphics[scale=0.38]{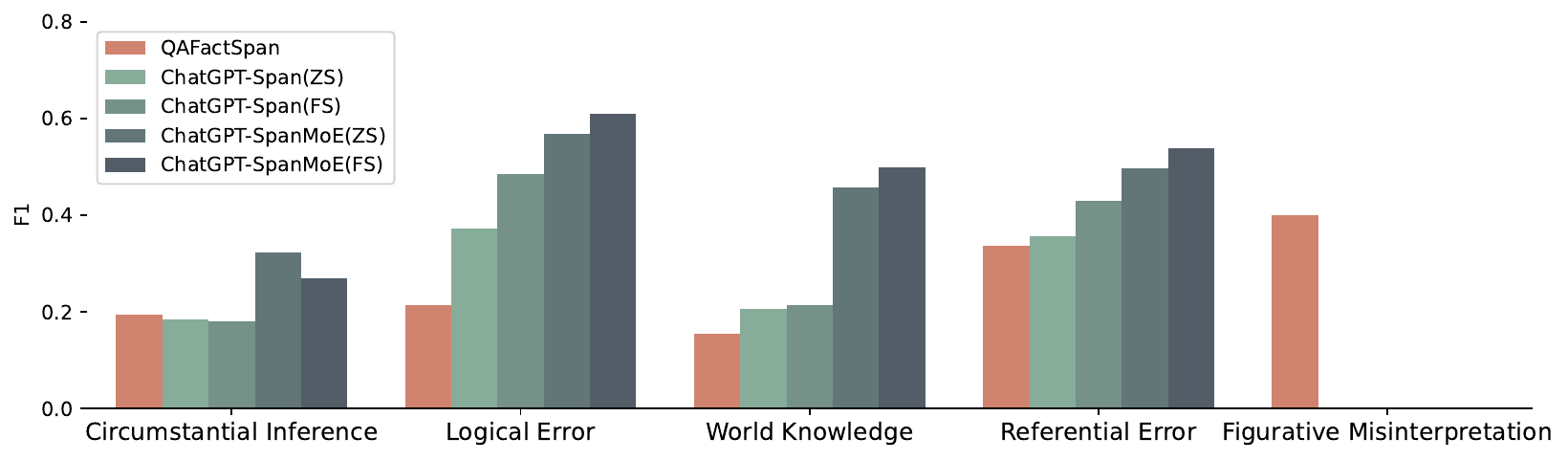}
    \caption{Span based F1 scores per error category}
    \label{fig:error_category_graph_f1}
\end{figure*}
\subsubsection{Binary Classification}
Binary classification metrics assess whether a summary is faithful or not 
by providing a single overarching score relative to the source. 

We incorporate four metrics into our evaluation framework: two question-answering-based metrics, namely \textbf{QAFactEval} \citep{fabbri2021qafacteval} and \textbf{QuestEval} \citep{scialom2021questeval}, alongside two natural language inference (NLI) based metrics, \textbf{SummaC-ZS} and \textbf{SummaC-Conv} \citep{laban2022summac}, which serve as our baseline measures. Both question-answering (QA) and natural language inference (NLI) metrics provide continuous scores for summarization. To translate these scores into binary factuality labels, we establish thresholds using a subset of 10\% of the evaluation data. Scores exceeding the designated threshold are classified as nonfactual. Distinct thresholds are determined for each metric and model type across all datasets. Further elaboration on this process can be found in Appendix \ref{sec:thresholding}.

Recent studies have also investigated the effectiveness of integrating ChatGPT prompts in error detection, demonstrating promising results compared to conventional metrics. Consequently, we include the established prompt \textbf{ChatGPT-Direct Assessment (ChatGPT-DA)} \citep{luo2023chatgpt} and evaluate its performance across both zero-shot and few-shot scenarios. The prompt is shown in Appendix \ref{subsec:chatgpt_da_prompt_details}.

\subsubsection{Span Detection}
\label{subsection:Span_Detection}
\subsubsection*{Baseline}
Span detection involves the meticulous identification of nonfactual or inconsistent spans within the summary when compared to the source document. As a standard baseline, we integrate \textbf{QAFactEval} and designate all spans labeled "unanswerable" by the metric as inconsistent. 

\subsubsection*{Our Approach}
In addition to the aforementioned baseline approach, we introduce two new prompt-based methodologies. These approaches are structured around two distinct subtasks.

a) Identification: This prompt focuses solely on extracting inconsistent spans based on provided definitions.
b) Verification: In this phase, sentences containing these nonfactual spans are compared with the source text using a prompt, that asks GPT to provide a consistency rating from 1 to 5 (detailed in Appendix \ref{subsec:verrification}). Spans are classified as nonfactual only if they receive a rating below 5 during verification.
Considering that summaries generated by LLMs tend to be more abstract and may involve inference, we aim to avoid extracting spans as nonfactual that rely solely on common sense, even if not explicitly supported by evidence. Our objective is to identify spans where information is either entirely fabricated or ambiguous based on the content. 

For our first approach, we use a generic prompt for identification (Appendix \ref{sec:chatgpt_span}) followed by verification and call this approach \textbf{ChatGPT-Span} .

To enhance the previous strategy, we introduce a "mixture of experts" concept for span identification (a). This method involves using distinct prompts for each error type outlined in our taxonomy (Section \ref{sec:taxonomy_err}). Each error type has a unique prompt (see Appendix \ref{sec:chatgpt_spanmoe}) given to GPT, allowing it to target each error type. After experts identify all error types, spans undergo verification (step 2) using the same procedure as before. 
This approach is called \textbf{ChatGPT-MoE}.

\begin{table*}
\centering

        \begin{tabular}{p{0.3\textwidth}p{0.075\textwidth}p{0.075\textwidth}p{0.075\textwidth}
        p{0.075\textwidth} p{0.075\textwidth}p{0.075\textwidth}}
        \toprule
         & \multicolumn{3}{c}{\textbf{  FT-Summ  }}  
         &\multicolumn{3}{c}{\textbf{  LLM  }} \\
         &\thead{F1} 
         &\thead{Precision}   
         &\thead{Recall} 
         &\thead{F1} 
         &\thead{Precision}   
         &\thead{Recall} \\
         \midrule
         QAFactEval & 10.16 & 7.55 & 15.52 & 8.54 & 7.49 & 9.91\\
         \midrule
         \multicolumn{6}{c}{Prompt-based (Our Approaches)}\\
         \midrule
         ChatGPT-Span(ZS) & 23.55 & 24.79 & 22.44 & 30.59 & 33.47 & 28.17\\
         ChatGPT-Span(FS) & 24.31 & 27.54 & 23.59 & 30.51 & 33.37 & 28.10\\
         ChatGPT-MoE (ZS) & 28.04 & 27.62 & 28.46 & 31.29 & 33.93 & 29.04  \\
         ChatGPT-MoE (FS) & \textbf{29.04} & \textbf{27.90} & \textbf{30.27} & \textbf{33.22} & \textbf{35.60} & \textbf{31.14}\\
         \bottomrule
         
        \end{tabular}
        \caption{F1-scores for fine-grained error detectors are shown for both FT-Summ and LLM models. The ChatGPT-MoE prompt metric exhibits superior performance, particularly in detecting circumstantial inference errors which leads to superior performance on LLMs compared to FT-Summ models. ZS and FS indicate Zero-
Shot and Few-Shot settings, respectively}
        \label{table:span_f1}
\end{table*}
\subsection{Evaluation Setup}
\subsubsection*{Metrics}
We use balanced accuracy as a metric for binary classification. It calculates the arithmetic mean of sensitivity and specificity, giving equal importance to minority and majority classes making it beneficial for imbalanced data. We use F-1 scores to compare predicted spans with annotated ones.

For binary classification, we also include span-level automated metrics. We predict a summary as consistent if no inconsistent spans are predicted, and inconsistent if at least one is predicted. In prompt-based few-shot setups, we use a four-shot approach, with two shots indicating inconsistency and two indicating consistency. 

\subsubsection*{Datasets and Models}
To compare metrics across benchmarks, we utilize all annotated FT-Summ models from Section \ref{sec:models} to compute balanced accuracy. It's important to note that the FacEval benchmark doesn't involve spans but focuses solely on error types. Therefore, our evaluation of span-level F1 scores on this benchmark is restricted to our LLM span annotations.
For assessing performance across error categories, we exclusively use summaries generated by BART, referred to as FT-Summ. 
Using our taxonomy, we classify annotated spans from previous benchmarks within BART summaries, enabling us to evaluate performance metrics on a per-category basis.\\
We also find that approximately 1\% of our dataset contains nonsensical errors, including grammatical and prompt inaccuracies. Since these errors mainly affect coherence rather than consistency, we have chosen to exclude this subset from our analysis.
\subsection{Results}

\subsubsection*{Binary Classification}
We start by discussing binary inconsistency detection results. Table \ref{tab:binary_metrics_results} presents balanced accuracy scores for FT-Summ and LLM models. 
Interestingly, prompt-based approaches, including direct assessment and span-based methods, perform less effectively in identifying LLM errors than FT-Summ errors (Figure \ref{fig:binary_classification_xformer_vs_llm}). Overall, prompt-based methods outperform standard QA and NLI metrics for both model types, though the improvement is less substantial for LLM models compared to FT-Summ models. We also show per category performance in Figure \ref{fig:error_category_graph_binary}.
\begin{table}[!htb]
    \centering
        \begin{tabular}{ccc}
        \toprule
         \textbf{Metric} &\textbf{FT-Summ} &\textbf{LLM} \\
         \midrule
         QuestEval & 47.54 & 49.47\\
         QAFactEval & 45.00 & 39.84\\
         SummaC-ZS & 43.29 & 49.70\\
         SummaC-Conv & 51.18 & 46.92\\
         ChatGPT-DA(ZS) & 73.00 & 60.34\\
         ChatGPT-DA(FS) & 72.06 & 61.61\\
         \midrule
         \multicolumn{3}{c}{Our Approaches}\\
         \midrule
         ChatGPT-Span(ZS) & 73.48 & 63.89\\
         ChatGPT-Span(FS) & 72.18 & 64.84\\
         ChatGPT-SpanMoE(ZS) & 73.77 & 67.96\\
         ChatGPT-SpanMoE(FS) & \textbf{75.61} & \textbf{70.27}\\
         \bottomrule
        \end{tabular}
        \caption{
Binary accuracy scores comparing factual label predictions of different metrics against human annotations. ChatGPT-Span and ChatGPT-SpanMoE outperform ChatGPT-DA and standard metrics, especially on LLM-generated summaries. ZS and FS indicate Zero-Shot and Few-Shot settings, respectively.}
        \label{tab:binary_metrics_results}
\end{table}
\subsubsection*{Span Detection}

Table \ref{table:span_f1} shows span-level F1 scores, comparing predicted spans with annotated ones. Results reveal that QAFactEval struggles with detecting inconsistent spans. In contrast, ChatGPT-Span demonstrates notably superior performance in both Zero-Shot(ZS) and Few-Shot scenarios(FS). With Few-Shot, it exhibits a 1-point increase in F1 scores for FT-Summ models, primarily enhancing precision. Moreover, incorporating ChatGPT-MoE yields a further improvement of nearly 5 points for FT-Summ and nearly 3 points for LLM summaries. Specifically, for FT-Summ, the most considerable enhancement lies in span recall, while for LLM models, the MoE prompt method affects both precision and recall.
When examining span based metrics per error-category, we observe that ChatGPT-MoE notably enhances the detection of Circumstantial Inference Errors as depicted in Figure \ref{fig:error_category_graph_f1}.
\section{Conclusion}

In summary, our work is the first to comprehensively assess Large Language Model (LLM) performance in dialogue summarization, revealing considerable inconsistencies that underscore the ongoing challenges in this area. Our findings emphasize the prevalence of circumstantial inferences, in summaries generated by GPT-4 and Alpaca-13b, indicating LLMs' proficiency in language understanding but their tendency to introduce conceptual inferences. Moreover, we demonstrate that existing metrics struggle to detect these nuanced errors effectively. Consequently, we advocate for performance evaluations on benchmarks utilizing newer models to better capture the capabilities and limitations of automatic metrics, given the evolving error distributions and types of newer LLMs compared to FT-Summ models. Furthermore, our incorporation of two prompt-based methods shows promising progress in identifying circumstantial inference errors, although further research is required to improve performance.
\section{Limitations and Ethics}
This study has limitations that should be noted. Firstly, the annotation process is resource-intensive and time-consuming. Consequently, we only benchmarked and annotated two Large Language Models (LLMs), which may not fully represent the behavior of all LLMs. Additionally, ethical considerations arise regarding the use of GPT-4 for prompt-based metrics. Being closed-source and expensive, its accessibility might be restricted, possibly widening the gap in research resources and impeding the reproducibility of our methodology.

\subsection{Citations}

\bibliography{anthology,custom}

\begin{thebibliography}{26}
\expandafter\ifx\csname natexlab\endcsname\relax\def\natexlab#1{#1}\fi

\bibitem[{Cao et~al.(2022)Cao, Dong, and Cheung}]{hallucinated_but_factual}
Meng Cao, Yue Dong, and Jackie Cheung. 2022.
\newblock \href {https://doi.org/10.18653/v1/2022.acl-long.236} {Hallucinated but factual! inspecting the factuality of hallucinations in abstractive summarization}.
\newblock In \emph{Proceedings of the 60th Annual Meeting of the Association for Computational Linguistics (Volume 1: Long Papers)}, pages 3340--3354, Dublin, Ireland. Association for Computational Linguistics.

\bibitem[{Chen and Yang(2020)}]{mvbart}
Jiaao Chen and Diyi Yang. 2020.
\newblock \href {https://doi.org/10.18653/v1/2020.emnlp-main.336} {Multi-view sequence-to-sequence models with conversational structure for abstractive dialogue summarization}.
\newblock In \emph{Proceedings of the 2020 Conference on Empirical Methods in Natural Language Processing (EMNLP)}, pages 4106--4118, Online. Association for Computational Linguistics.

\bibitem[{Chen et~al.(2021)Chen, Liu, Chen, and Zhang}]{chen2021dialogsum}
Yulong Chen, Yang Liu, Liang Chen, and Yue Zhang. 2021.
\newblock Dialogsum: A real-life scenario dialogue summarization dataset.
\newblock \emph{arXiv preprint arXiv:2105.06762}.

\bibitem[{Dong et~al.(2019)Dong, Yang, Wang, Wei, Liu, Wang, Gao, Zhou, and Hon}]{unilm}
Li~Dong, Nan Yang, Wenhui Wang, Furu Wei, Xiaodong Liu, Yu~Wang, Jianfeng Gao, Ming Zhou, and Hsiao-Wuen Hon. 2019.
\newblock Unified language model pre-training for natural language understanding and generation.
\newblock \emph{Advances in neural information processing systems}, 32.

\bibitem[{Fabbri et~al.(2021)Fabbri, Wu, Liu, and Xiong}]{fabbri2021qafacteval}
Alexander~R Fabbri, Chien-Sheng Wu, Wenhao Liu, and Caiming Xiong. 2021.
\newblock Qafacteval: Improved qa-based factual consistency evaluation for summarization.
\newblock \emph{arXiv preprint arXiv:2112.08542}.

\bibitem[{Gao et~al.(2023)Gao, Wan, Su, Wang, and Huai}]{gao2023reference}
Mingqi Gao, Xiaojun Wan, Jia Su, Zhefeng Wang, and Baoxing Huai. 2023.
\newblock Reference matters: Benchmarking factual error correction for dialogue summarization with fine-grained evaluation framework.
\newblock \emph{arXiv preprint arXiv:2306.05119}.

\bibitem[{Gliwa et~al.(2019)Gliwa, Mochol, Biesek, and Wawer}]{gliwa2019samsum}
Bogdan Gliwa, Iwona Mochol, Maciej Biesek, and Aleksander Wawer. 2019.
\newblock Samsum corpus: A human-annotated dialogue dataset for abstractive summarization.
\newblock \emph{arXiv preprint arXiv:1911.12237}.

\bibitem[{Goyal and Durrett(2021)}]{dae}
Tanya Goyal and Greg Durrett. 2021.
\newblock \href {https://doi.org/10.18653/v1/2021.naacl-main.114} {Annotating and modeling fine-grained factuality in summarization}.
\newblock In \emph{Proceedings of the 2021 Conference of the North American Chapter of the Association for Computational Linguistics: Human Language Technologies}, pages 1449--1462, Online. Association for Computational Linguistics.

\bibitem[{Goyal et~al.(2022)Goyal, Li, and Durrett}]{goyal2022news}
Tanya Goyal, Junyi~Jessy Li, and Greg Durrett. 2022.
\newblock News summarization and evaluation in the era of gpt-3.
\newblock \emph{arXiv preprint arXiv:2209.12356}.

\bibitem[{Grice(1975)}]{grice1975logic}
Herbert~P Grice. 1975.
\newblock Logic and conversation.
\newblock In \emph{Speech acts}, pages 41--58. Brill.

\bibitem[{Huang et~al.(2020)Huang, Cui, Yang, Bao, Wang, Xie, and Zhang}]{polytype}
Dandan Huang, Leyang Cui, Sen Yang, Guangsheng Bao, Kun Wang, Jun Xie, and Yue Zhang. 2020.
\newblock \href {https://doi.org/10.18653/v1/2020.emnlp-main.33} {What have we achieved on text summarization?}
\newblock In \emph{Proceedings of the 2020 Conference on Empirical Methods in Natural Language Processing (EMNLP)}, pages 446--469, Online. Association for Computational Linguistics.

\bibitem[{Laban et~al.(2022)Laban, Schnabel, Bennett, and Hearst}]{laban2022summac}
Philippe Laban, Tobias Schnabel, Paul~N Bennett, and Marti~A Hearst. 2022.
\newblock Summac: Re-visiting nli-based models for inconsistency detection in summarization.
\newblock \emph{Transactions of the Association for Computational Linguistics}, 10:163--177.

\bibitem[{Lewis et~al.(2020)Lewis, Liu, Goyal, Ghazvininejad, Mohamed, Levy, Stoyanov, and Zettlemoyer}]{lewis-etal-2020-bart}
Mike Lewis, Yinhan Liu, Naman Goyal, Marjan Ghazvininejad, Abdelrahman Mohamed, Omer Levy, Veselin Stoyanov, and Luke Zettlemoyer. 2020.
\newblock \href {https://doi.org/10.18653/v1/2020.acl-main.703} {{BART}: Denoising sequence-to-sequence pre-training for natural language generation, translation, and comprehension}.
\newblock In \emph{Proceedings of the 58th Annual Meeting of the Association for Computational Linguistics}, pages 7871--7880, Online. Association for Computational Linguistics.

\bibitem[{Liu et~al.(2021{\natexlab{a}})Liu, Zou, Zhang, Chen, Ding, Yuan, and Wang}]{condigsum}
Junpeng Liu, Yanyan Zou, Hainan Zhang, Hongshen Chen, Zhuoye Ding, Caixia Yuan, and Xiaojie Wang. 2021{\natexlab{a}}.
\newblock \href {https://doi.org/10.18653/v1/2021.findings-emnlp.106} {Topic-aware contrastive learning for abstractive dialogue summarization}.
\newblock In \emph{Findings of the Association for Computational Linguistics: EMNLP 2021}, pages 1229--1243, Punta Cana, Dominican Republic. Association for Computational Linguistics.

\bibitem[{Liu et~al.(2021{\natexlab{b}})Liu, Shi, and Chen}]{corefbart}
Zhengyuan Liu, Ke~Shi, and Nancy Chen. 2021{\natexlab{b}}.
\newblock \href {https://doi.org/10.18653/v1/2021.sigdial-1.53} {Coreference-aware dialogue summarization}.
\newblock In \emph{Proceedings of the 22nd Annual Meeting of the Special Interest Group on Discourse and Dialogue}, pages 509--519, Singapore and Online. Association for Computational Linguistics.

\bibitem[{Luo et~al.(2023)Luo, Xie, and Ananiadou}]{luo2023chatgpt}
Zheheng Luo, Qianqian Xie, and Sophia Ananiadou. 2023.
\newblock Chatgpt as a factual inconsistency evaluator for abstractive text summarization.
\newblock \emph{arXiv preprint arXiv:2303.15621}.

\bibitem[{Maynez et~al.(2020)Maynez, Narayan, Bohnet, and McDonald}]{xsumfaith}
Joshua Maynez, Shashi Narayan, Bernd Bohnet, and Ryan McDonald. 2020.
\newblock \href {https://doi.org/10.18653/v1/2020.acl-main.173} {On faithfulness and factuality in abstractive summarization}.
\newblock In \emph{Proceedings of the 58th Annual Meeting of the Association for Computational Linguistics}, pages 1906--1919, Online. Association for Computational Linguistics.

\bibitem[{OpenAI et~al.(2023)OpenAI, :, Achiam, Adler, Agarwal, Ahmad, Akkaya, Aleman, Almeida, Altenschmidt, Altman, Anadkat, Avila, Babuschkin, Balaji, Balcom, Baltescu, Bao, Bavarian, Belgum, Bello, Berdine, Bernadett-Shapiro, Berner, Bogdonoff, Boiko, Boyd, Brakman, Brockman, Brooks, Brundage, Button, Cai, Campbell, Cann, Carey, Carlson, Carmichael, Chan, Chang, Chantzis, Chen, Chen, Chen, Chen, Chen, Chess, Cho, Chu, Chung, Cummings, Currier, Dai, Decareaux, Degry, Deutsch, Deville, Dhar, Dohan, Dowling, Dunning, Ecoffet, Eleti, Eloundou, Farhi, Fedus, Felix, Fishman, Forte, Fulford, Gao, Georges, Gibson, Goel, Gogineni, Goh, Gontijo-Lopes, Gordon, Grafstein, Gray, Greene, Gross, Gu, Guo, Hallacy, Han, Harris, He, Heaton, Heidecke, Hesse, Hickey, Hickey, Hoeschele, Houghton, Hsu, Hu, Hu, Huizinga, Jain, Jain, Jang, Jiang, Jiang, Jin, Jin, Jomoto, Jonn, Jun, Kaftan, Łukasz Kaiser, Kamali, Kanitscheider, Keskar, Khan, Kilpatrick, Kim, Kim, Kim, Kirchner, Kiros, Knight, Kokotajlo, Łukasz Kondraciuk,
  Kondrich, Konstantinidis, Kosic, Krueger, Kuo, Lampe, Lan, Lee, Leike, Leung, Levy, Li, Lim, Lin, Lin, Litwin, Lopez, Lowe, Lue, Makanju, Malfacini, Manning, Markov, Markovski, Martin, Mayer, Mayne, McGrew, McKinney, McLeavey, McMillan, McNeil, Medina, Mehta, Menick, Metz, Mishchenko, Mishkin, Monaco, Morikawa, Mossing, Mu, Murati, Murk, Mély, Nair, Nakano, Nayak, Neelakantan, Ngo, Noh, Ouyang, O'Keefe, Pachocki, Paino, Palermo, Pantuliano, Parascandolo, Parish, Parparita, Passos, Pavlov, Peng, Perelman, de~Avila Belbute~Peres, Petrov, de~Oliveira~Pinto, Michael, Pokorny, Pokrass, Pong, Powell, Power, Power, Proehl, Puri, Radford, Rae, Ramesh, Raymond, Real, Rimbach, Ross, Rotsted, Roussez, Ryder, Saltarelli, Sanders, Santurkar, Sastry, Schmidt, Schnurr, Schulman, Selsam, Sheppard, Sherbakov, Shieh, Shoker, Shyam, Sidor, Sigler, Simens, Sitkin, Slama, Sohl, Sokolowsky, Song, Staudacher, Such, Summers, Sutskever, Tang, Tezak, Thompson, Tillet, Tootoonchian, Tseng, Tuggle, Turley, Tworek, Uribe, Vallone,
  Vijayvergiya, Voss, Wainwright, Wang, Wang, Wang, Ward, Wei, Weinmann, Welihinda, Welinder, Weng, Weng, Wiethoff, Willner, Winter, Wolrich, Wong, Workman, Wu, Wu, Wu, Xiao, Xu, Yoo, Yu, Yuan, Zaremba, Zellers, Zhang, Zhang, Zhao, Zheng, Zhuang, Zhuk, and Zoph}]{openai2023gpt4}
OpenAI, :, Josh Achiam, Steven Adler, Sandhini Agarwal, Lama Ahmad, Ilge Akkaya, Florencia~Leoni Aleman, Diogo Almeida, Janko Altenschmidt, Sam Altman, Shyamal Anadkat, Red Avila, Igor Babuschkin, Suchir Balaji, Valerie Balcom, Paul Baltescu, Haiming Bao, Mo~Bavarian, Jeff Belgum, Irwan Bello, Jake Berdine, Gabriel Bernadett-Shapiro, Christopher Berner, Lenny Bogdonoff, Oleg Boiko, Madelaine Boyd, Anna-Luisa Brakman, Greg Brockman, Tim Brooks, Miles Brundage, Kevin Button, Trevor Cai, Rosie Campbell, Andrew Cann, Brittany Carey, Chelsea Carlson, Rory Carmichael, Brooke Chan, Che Chang, Fotis Chantzis, Derek Chen, Sully Chen, Ruby Chen, Jason Chen, Mark Chen, Ben Chess, Chester Cho, Casey Chu, Hyung~Won Chung, Dave Cummings, Jeremiah Currier, Yunxing Dai, Cory Decareaux, Thomas Degry, Noah Deutsch, Damien Deville, Arka Dhar, David Dohan, Steve Dowling, Sheila Dunning, Adrien Ecoffet, Atty Eleti, Tyna Eloundou, David Farhi, Liam Fedus, Niko Felix, Simón~Posada Fishman, Juston Forte, Isabella Fulford, Leo Gao,
  Elie Georges, Christian Gibson, Vik Goel, Tarun Gogineni, Gabriel Goh, Rapha Gontijo-Lopes, Jonathan Gordon, Morgan Grafstein, Scott Gray, Ryan Greene, Joshua Gross, Shixiang~Shane Gu, Yufei Guo, Chris Hallacy, Jesse Han, Jeff Harris, Yuchen He, Mike Heaton, Johannes Heidecke, Chris Hesse, Alan Hickey, Wade Hickey, Peter Hoeschele, Brandon Houghton, Kenny Hsu, Shengli Hu, Xin Hu, Joost Huizinga, Shantanu Jain, Shawn Jain, Joanne Jang, Angela Jiang, Roger Jiang, Haozhun Jin, Denny Jin, Shino Jomoto, Billie Jonn, Heewoo Jun, Tomer Kaftan, Łukasz Kaiser, Ali Kamali, Ingmar Kanitscheider, Nitish~Shirish Keskar, Tabarak Khan, Logan Kilpatrick, Jong~Wook Kim, Christina Kim, Yongjik Kim, Hendrik Kirchner, Jamie Kiros, Matt Knight, Daniel Kokotajlo, Łukasz Kondraciuk, Andrew Kondrich, Aris Konstantinidis, Kyle Kosic, Gretchen Krueger, Vishal Kuo, Michael Lampe, Ikai Lan, Teddy Lee, Jan Leike, Jade Leung, Daniel Levy, Chak~Ming Li, Rachel Lim, Molly Lin, Stephanie Lin, Mateusz Litwin, Theresa Lopez, Ryan Lowe,
  Patricia Lue, Anna Makanju, Kim Malfacini, Sam Manning, Todor Markov, Yaniv Markovski, Bianca Martin, Katie Mayer, Andrew Mayne, Bob McGrew, Scott~Mayer McKinney, Christine McLeavey, Paul McMillan, Jake McNeil, David Medina, Aalok Mehta, Jacob Menick, Luke Metz, Andrey Mishchenko, Pamela Mishkin, Vinnie Monaco, Evan Morikawa, Daniel Mossing, Tong Mu, Mira Murati, Oleg Murk, David Mély, Ashvin Nair, Reiichiro Nakano, Rajeev Nayak, Arvind Neelakantan, Richard Ngo, Hyeonwoo Noh, Long Ouyang, Cullen O'Keefe, Jakub Pachocki, Alex Paino, Joe Palermo, Ashley Pantuliano, Giambattista Parascandolo, Joel Parish, Emy Parparita, Alex Passos, Mikhail Pavlov, Andrew Peng, Adam Perelman, Filipe de~Avila Belbute~Peres, Michael Petrov, Henrique~Ponde de~Oliveira~Pinto, Michael, Pokorny, Michelle Pokrass, Vitchyr Pong, Tolly Powell, Alethea Power, Boris Power, Elizabeth Proehl, Raul Puri, Alec Radford, Jack Rae, Aditya Ramesh, Cameron Raymond, Francis Real, Kendra Rimbach, Carl Ross, Bob Rotsted, Henri Roussez, Nick Ryder,
  Mario Saltarelli, Ted Sanders, Shibani Santurkar, Girish Sastry, Heather Schmidt, David Schnurr, John Schulman, Daniel Selsam, Kyla Sheppard, Toki Sherbakov, Jessica Shieh, Sarah Shoker, Pranav Shyam, Szymon Sidor, Eric Sigler, Maddie Simens, Jordan Sitkin, Katarina Slama, Ian Sohl, Benjamin Sokolowsky, Yang Song, Natalie Staudacher, Felipe~Petroski Such, Natalie Summers, Ilya Sutskever, Jie Tang, Nikolas Tezak, Madeleine Thompson, Phil Tillet, Amin Tootoonchian, Elizabeth Tseng, Preston Tuggle, Nick Turley, Jerry Tworek, Juan Felipe~Cerón Uribe, Andrea Vallone, Arun Vijayvergiya, Chelsea Voss, Carroll Wainwright, Justin~Jay Wang, Alvin Wang, Ben Wang, Jonathan Ward, Jason Wei, CJ~Weinmann, Akila Welihinda, Peter Welinder, Jiayi Weng, Lilian Weng, Matt Wiethoff, Dave Willner, Clemens Winter, Samuel Wolrich, Hannah Wong, Lauren Workman, Sherwin Wu, Jeff Wu, Michael Wu, Kai Xiao, Tao Xu, Sarah Yoo, Kevin Yu, Qiming Yuan, Wojciech Zaremba, Rowan Zellers, Chong Zhang, Marvin Zhang, Shengjia Zhao, Tianhao
  Zheng, Juntang Zhuang, William Zhuk, and Barret Zoph. 2023.
\newblock \href {http://arxiv.org/abs/2303.08774} {Gpt-4 technical report}.

\bibitem[{Scialom et~al.(2021)Scialom, Dray, Gallinari, Lamprier, Piwowarski, Staiano, and Wang}]{scialom2021questeval}
Thomas Scialom, Paul-Alexis Dray, Patrick Gallinari, Sylvain Lamprier, Benjamin Piwowarski, Jacopo Staiano, and Alex Wang. 2021.
\newblock Questeval: Summarization asks for fact-based evaluation.
\newblock \emph{arXiv preprint arXiv:2103.12693}.

\bibitem[{Tang et~al.(2022)Tang, Goyal, Fabbri, Laban, Xu, Yavuz, Kry{\'s}ci{\'n}ski, Rousseau, and Durrett}]{tang2022understanding}
Liyan Tang, Tanya Goyal, Alexander~R Fabbri, Philippe Laban, Jiacheng Xu, Semih Yavuz, Wojciech Kry{\'s}ci{\'n}ski, Justin~F Rousseau, and Greg Durrett. 2022.
\newblock Understanding factual errors in summarization: Errors, summarizers, datasets, error detectors.
\newblock \emph{arXiv preprint arXiv:2205.12854}.

\bibitem[{Taori et~al.(2023)Taori, Gulrajani, Zhang, Dubois, Li, Guestrin, Liang, and Hashimoto}]{alpaca}
Rohan Taori, Ishaan Gulrajani, Tianyi Zhang, Yann Dubois, Xuechen Li, Carlos Guestrin, Percy Liang, and Tatsunori~B. Hashimoto. 2023.
\newblock Stanford alpaca: An instruction-following llama model.
\newblock \url{https://github.com/tatsu-lab/stanford_alpaca}.

\bibitem[{Wang et~al.(2022)Wang, Zhang, Zhang, Chen, and Li}]{wang2022analyzing}
Bin Wang, Chen Zhang, Yan Zhang, Yiming Chen, and Haizhou Li. 2022.
\newblock Analyzing and evaluating faithfulness in dialogue summarization.
\newblock \emph{arXiv preprint arXiv:2210.11777}.

\bibitem[{Wu et~al.(2021)Wu, Liu, Liu, Stenetorp, and Xiong}]{cods}
Chien-Sheng Wu, Linqing Liu, Wenhao Liu, Pontus Stenetorp, and Caiming Xiong. 2021.
\newblock \href {https://doi.org/10.18653/v1/2021.findings-acl.454} {Controllable abstractive dialogue summarization with sketch supervision}.
\newblock In \emph{Findings of the Association for Computational Linguistics: ACL-IJCNLP 2021}, pages 5108--5122, Online. Association for Computational Linguistics.

\bibitem[{Yang et~al.(2023)Yang, Li, Zhang, Chen, and Cheng}]{yang2023exploring}
Xianjun Yang, Yan Li, Xinlu Zhang, Haifeng Chen, and Wei Cheng. 2023.
\newblock Exploring the limits of chatgpt for query or aspect-based text summarization.
\newblock \emph{arXiv preprint arXiv:2302.08081}.

\bibitem[{Zhang et~al.(2023)Zhang, Ladhak, Durmus, Liang, McKeown, and Hashimoto}]{zhang2023benchmarking}
Tianyi Zhang, Faisal Ladhak, Esin Durmus, Percy Liang, Kathleen McKeown, and Tatsunori~B Hashimoto. 2023.
\newblock Benchmarking large language models for news summarization.
\newblock \emph{arXiv preprint arXiv:2301.13848}.

\bibitem[{Zhu et~al.(2023)Zhu, Qi, and Lau}]{zhu2023annotating}
Rongxin Zhu, Jianzhong Qi, and Jey~Han Lau. 2023.
\newblock Annotating and detecting fine-grained factual errors for dialogue summarization.
\newblock \emph{arXiv preprint arXiv:2305.16548}.

\end{thebibliography}
\bibliographystyle{acl_natbib}

\appendix

\section{Annotation}
\label{sec:annotation}

\subsection{Annotator Recruitment}
We hired annotators through UpWork. Candidates underwent a qualifying round and an interview where they had to explain marked errors. Ultimately, we selected two expert proofreaders who were paid \$18USD and \$22 USD per hour, respectively.

\subsection{Annotator Instruction}
\begin{figure*}
    \centering
    \includegraphics[scale=0.45]{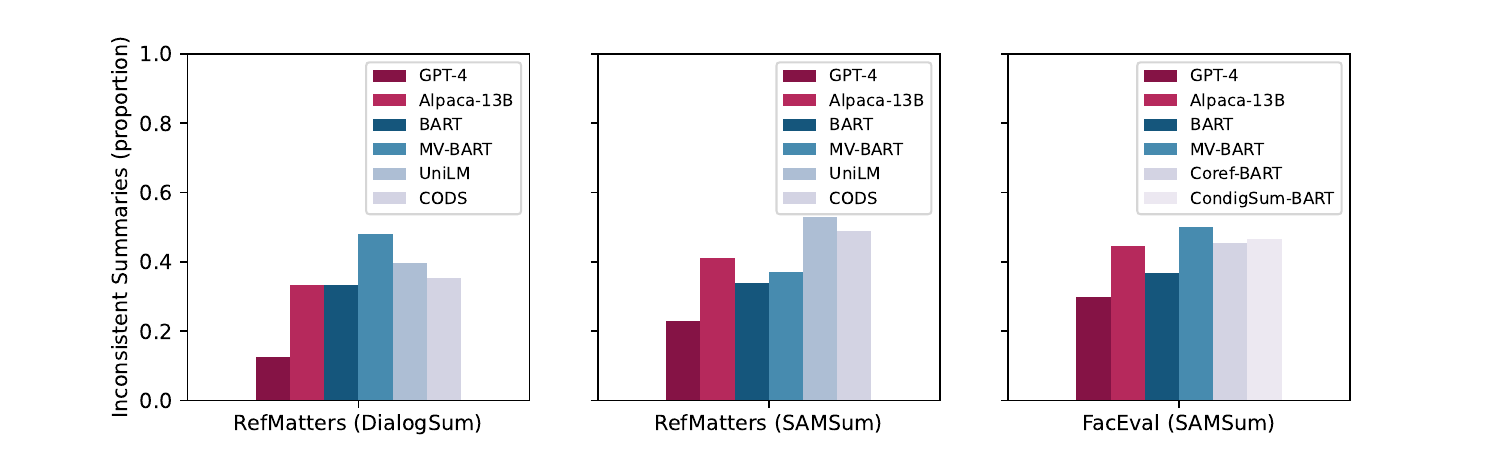}
    \caption{Inconsistency rate of all models per dataset. We see that Alpaca-13B is competitive with BART with respect to consistency on the DialogSum dataset but is outperformed on the SAMSum datasets}
\label{fig:llm_vs_transfomrers_error_rates_dataset_models}
\end{figure*}

The following were the instructions provided to annotators to mark spans as inconsistent. 

\textit{Identify minimal spans in the summary that:\\
a) Misrepresent information from the source: If a span contradicts or distorts information with respect to the source, annotate the evidence sentences from the source that demonstrate this inconsistency and select the span as inconsistent.\\
b) Introduce new information not supported by evidence in the source: If the summary includes new information that is neither common knowledge nor a logical inference but relies on external facts or deductions, mark these spans as inconsistent. In this case, evidence sentences may not be available for annotation.}

\section{Error Rate per Dataset}
\begin{figure*}
    \centering
    \includegraphics[scale=0.45]{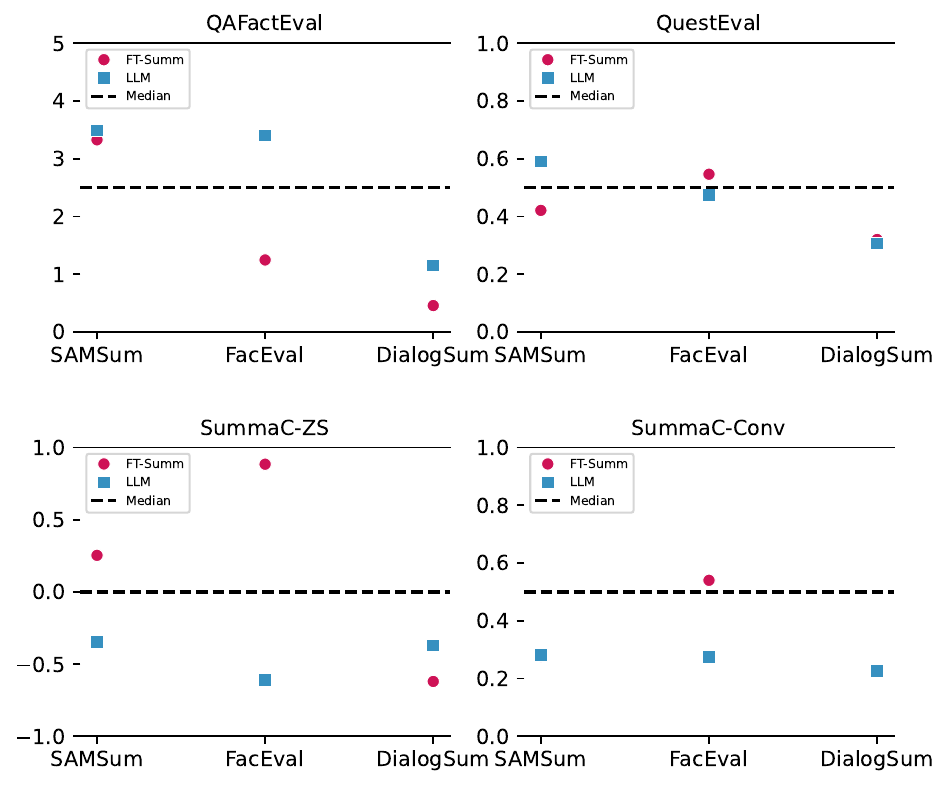}
    \caption{Thresholds are displayed for each metric on each dataset and model category.}
\label{fig:thresholds}
\end{figure*}
\label{sec:error_rate_dataset}
In figure \ref{fig:llm_vs_transfomrers_error_rates_dataset_models} we provide the inconsistency rates for all models across each dataset. GPT-4 exhibits the highest consistency across all datasets. However, Alpaca-13b shows similar performance to BART on the dialogsum dataset but is surpassed by BART on the SAMSum datasets.

\section{Thresholding for Binary Classification}
\label{sec:thresholding}
We use a subset of the evaluation data and apply thresholding to convert continuous scores into binary labels. This subset comprises approximately 150 source-summary pairs. The thresholds are individually determined for each metric, dataset, and model category.
The thresholds are displayed in Figure \ref{fig:thresholds}

\section{Prompt Details}
\label{sec:prompt_details}

\subsection{ChatGPT-Direct Assessment}
\label{subsec:chatgpt_da_prompt_details}


\begin{tcolorbox}[colback=gray!10, colframe=gray!20, breakable]
Decide if the Summary is consistent with the corresponding Content. Note that consistency means all information in the summary is supported by the Content.
Answer "yes" for consistent and "no" for inconsistent: \\
Content: $\{\{Dialogue\}\}$\\
Summary: $\{\{Summary\}\}$\\
Answer
\end{tcolorbox}

\subsection{ChatGPT-Span}
\label{sec:chatgpt_span}

\subsubsection{Identification}
\begin{tcolorbox}[colback=gray!10, colframe=gray!20, breakable]
Identify and list spans in the summary which are not supported by evidence from the content; if there are no unsupported spans, respond with "None" \\
Content: $\{\{Dialogue\}\}$\\
Summary: $\{\{Summary\}\}$\\
Answer
\end{tcolorbox}

\subsubsection{Verification}
\label{subsec:verrification}
\begin{tcolorbox}[colback=gray!10, colframe=gray!20, breakable]
Content: $\{\{Dialogue\}\}$\\
Assess the extent to which the specified span in the following sentence is supported by evidence from the content, using a scale of 1 to 5, where 1 indicates no supporting evidence and 5 indicates full support from the evidence provided within the content \\
Span: $\{\{Span\}\}$\\
Sentence: $\{\{Summary Sentence\}\}$\\
Answer:
\end{tcolorbox}

\subsection{ChatGPT-SpanMoE}
\label{sec:chatgpt_spanmoe}

\subsubsection{Identification}
\textbf{Circumstantial Inference}
\begin{tcolorbox}[colback=gray!10, colframe=gray!20, breakable]
Error Definition: Circumstantial inference in summaries 
is inferred supplemental information, 
not explicitly stated in the content 
but derived from circumstantial evidence, 
often intentionally omitted in the content 
and assumed to be shared knowledge among participants 
in adherence to the principle 
of providing sufficient information 
without unnecessary details. \\
Task Definition: Extract spans from the summary
that are circumstantial inferences.\\
Ensure the spans are the minimal erroneous spans. 
List each span in a new line; 
if there are no such spans respond with None
Content: $\{\{Dialogue\}\}$\\
Summary: $\{\{Summary\}\}$\\
Answer
\end{tcolorbox}

\textbf{Logical Error}
\begin{tcolorbox}[colback=gray!10, colframe=gray!20, breakable]
Error Definition: Logical inference errors in summaries arise from drawing conclusions or making deductions that deviate from the logical flow of content, leading to inaccuracies or misunderstandings in the representation of information or ideas. \\
Task Definition: Extract spans from the summary that are logical errors.\\
Ensure the spans are the minimal erroneous spans. List each span in a new line; if there are no such spans respond with None
Content: $\{\{Dialogue\}\}$\\
Summary: $\{\{Summary\}\}$\\
Answer
\end{tcolorbox}

\textbf{World Knowledge}
\begin{tcolorbox}[colback=gray!10, colframe=gray!20, breakable]
Error Definition: Factual extrapolations are real-world facts added in a summary,
not explicitly mentioned in the original conversation. \\
Task Definition: Extract spans from the summary that introduces additional details, 
constituting general facts or world knowledge not explicitly stated in the original content.\\
Ensure the spans are the minimal erroneous spans. 
List each span in a new line; if there are no such spans respond with None
Content: $\{\{Dialogue\}\}$\\
Summary: $\{\{Summary\}\}$\\
Answer
\end{tcolorbox}

\textbf{Referential Error}
\begin{tcolorbox}[colback=gray!10, colframe=gray!20, breakable]
Error Definition: To identify referential errors, check for inconsistencies with respect to the content in linking pronouns, terms, or entities to their correct referents. Also look for instances of misattributions where statements or actions are inaccurately assigned to the wrong speaker or participant, resulting in content representation inaccuracies. \\
Task Definition: Extract spans from the summary that are referential errors.\\
Ensure the spans are the minimal erroneous spans. List each span in a new line; if there are no such spans respond with None
Content: $\{\{Dialogue\}\}$\\
Summary: $\{\{Summary\}\}$\\
Answer
\end{tcolorbox}

\textbf{Figurative Error}
\begin{tcolorbox}[colback=gray!10, colframe=gray!20, breakable]
Error Definition: Figurative misrepresentation occurs when non-literal information in the content is inaccurately portrayed or misunderstood as literal statements in the summary, distorting the intended meaning or message. \\
Task Definition: Extract spans from the summary that figuratively misrepresent information in the content.\\
Ensure the spans are the minimal erroneous spans. List each span in a new line; if there are no such spans respond with None
Content: $\{\{Dialogue\}\}$\\
Summary: $\{\{Summary\}\}$\\
Answer
\end{tcolorbox}

The verification step that follows the spans extracted from the above prompts is the same as displayed in \ref{subsec:verrification}
\end{document}